\documentclass[lettersize,journal]{IEEEtran}
\usepackage{amsmath,amsfonts}
\usepackage{array}
\usepackage[caption=false,font=normalsize,labelfont=sf,textfont=sf]{subfig}
\usepackage[colorlinks=true, linkcolor=blue, citecolor=blue]{hyperref}
\usepackage{textcomp}
\usepackage{stfloats}
\usepackage{url}
\usepackage{verbatim}
\usepackage{graphicx}
\usepackage{multirow}
\usepackage{cite}
\usepackage{caption}
\usepackage{float}
\usepackage{placeins}
\usepackage{balance}
\usepackage{algorithm}
\usepackage[noend]{algpseudocode}
\usepackage{xcolor}

\begin{document}
\definecolor{commentclr}{RGB}{140,140,140}

\title{Variational Network with Wavelet-based UNET in
Accelerated MRI Reconstruction from Under Sampled K-space Data}

\author{\uppercase{Yasir Arafat Prodhan}, \IEEEmembership{Member, IEEE},
\uppercase{Dr. Shaikh Anowarul Fattah}
\IEEEmembership{Senior Member, IEEE}

\thanks{This paper was produced by the IEEE Publication Technology Group. They are in Piscataway, NJ.}
\thanks{Manuscript received April 19, 2021; revised August 16, 2021.}}

\markboth{Journal of \LaTeX\ Class Files,~Vol.~14, No.~8, August~2021}%
{Shell \MakeLowercase{\textit{et al.}}: A Sample Article Using IEEEtran.cls for IEEE Journals}


\maketitle

\begin{abstract}
Fully sampled MRI acquisitions require dense traversal of k-space, resulting in inherently long scan times that limit clinical throughput and increase susceptibility to patient motion. A common strategy to accelerate MRI is to acquire only a subset of k-space measurements and reconstruct the missing information computationally. However, direct reconstruction from undersampled k-space is a severely ill-posed inverse problem that introduces aliasing artifacts, noise amplification, and loss of fine anatomical details. Conventional methods such as Parallel Imaging (PI) and Compressed Sensing (CS) partially address these challenges, while recent deep learning approaches have significantly improved reconstruction quality. Nevertheless, preserving high-frequency structures and maintaining robustness under aggressive undersampling remain challenging for existing convolution-based variational network frameworks. In this work, we propose a Variational Network with Wavelet-based U-Net (W-UNet) for accelerated MRI reconstruction from undersampled k-space data. The proposed framework combines physics-guided iterative reconstruction with learnable multi-scale frequency representations. Specifically, standard pooling operations are replaced by Discrete Wavelet Transform (DWT) and Inverse Wavelet Transform (IDWT) modules, enabling lossless downsampling while retaining both low-frequency structural information and high-frequency edge details. Integrated within the refinement and sensitivity map estimation stages of a variational network, the proposed design improves artifact suppression, feature preservation, and reconstruction fidelity for both single-coil and multi-coil settings. Our method achieves state-of-the-art results on the fastMRI knee and M4Raw brain datasets. We further validate the effectiveness of wavelet-based feature decomposition through comprehensive ablation studies, confirming its contribution to accelerated MRI reconstruction.
\end{abstract}

\begin{IEEEkeywords}
Variational Network, Wavelet-based CNN, MRI Reconstruction, Accelerated MRI, Deep Learning, Image Quality, SSIM, PSNR
\end{IEEEkeywords}

\section{Introduction}

Magnetic Resonance Imaging (MRI) is a widely used non-invasive imaging modality 
that provides high-resolution visualization of anatomical structures and soft 
tissues. Despite its diagnostic superiority over modalities such as 
Computed Tomography (CT) and Positron Emission Tomography (PET), \textbf{MRI}\cite{HAMILTON201771} suffers 
from inherently long acquisition times. This limitation arises from the 
sequential sampling of the spatial-frequency domain, known as k-space, where each 
line is acquired individually. As a result, fully sampled MRI acquisition is 
time-consuming, increases sensitivity to patient motion, raises clinical costs, 
and limits imaging throughput.

To accelerate MRI acquisition, undersampling strategies are employed in k-space, 
where only a subset of measurements is collected. However, undersampling violates 
the Nyquist sampling criterion, resulting in an ill-posed inverse problem. 
Direct reconstruction from undersampled k-space introduces aliasing artifacts, 
noise amplification, and loss of fine anatomical details. Therefore, the central 
challenge in accelerated MRI reconstruction is to recover high-fidelity images 
from incomplete measurements while preserving structural integrity.\\

Early approaches to accelerate MRI acquisition include Parallel Imaging (\textbf{PI}) and Compressed Sensing (\textbf{CS})\cite{Jeromin2012OptimalCSfMRI, 10.1109/TIT.2006.871582}. In PI, 
methods such as Sensitivity Encoding \textbf{SENSE})\cite{Pruessmann1999SENSESE}
and Generalized Autocalibrating Partially Parallel Acquisitions (\textbf{GRAPPA})\cite{https://doi.org/10.1002/mrm.10171} exploit multi-coil spatial sensitivity information to 
reconstruct missing k-space data. Low-rank structured parallel imaging reconstruction method \textbf{P-LORAKS}\cite{6678771} were later introduced to exploit redundancies in local k-space neighborhoods by jointly enforcing intra-channel and inter-channel k-space constraints.
As reviewed in \textbf{Larkman and Nunes}\cite{Larkman2007ParallelMRI}, PI methods can suffer from signal-to-noise ratio degradation and noise amplification, particularly at 
higher acceleration factors due to g-factor penalties. Meanwhile, although CS enables faster imaging through sparse reconstruction, 
\textbf{Sharma et al.}\cite{Sharma2013ClinicalImageQualityCSMRI}- Clinical 
CS Neuroimaging Assessment reported ringing artifacts and image blurring that may reduce diagnostic image quality at higher accelerations. Compressed Sensing regularizes the reconstruction using sparsity priors in transform domains effective for moderate undersampling, but their performance degrade at higher acceleration factors. These limitations have driven the development of deep learning approaches, where neural networks learn data-driven priors and nonlinear mappings from undersampled measurements.\\

Data-driven approaches can be categorized according to the domain in which 
learning is performed. Image-domain methods remove aliasing artifacts after 
transforming undersampled k-space into image space. Networks such as 
\textbf{DAGAN} \cite{8233175}, \textbf{DeepComplexMRI}\cite{WANG2020136}, and 
\textbf{RefineGAN}\cite{8327637} demonstrate strong reconstruction performance. \textbf{OUCR}\cite{10.1007/978-3-030-87231-1_2} further improved image-domain reconstruction by combining overcomplete and undercomplete convolutional recurrent branches to jointly preserve fine local structures and global contextual information. However, as these methods operate on already degraded images, high-frequency details may be partially lost.

K-space domain methods, such as \textbf{RAKI}\cite{https://doi.org/10.1002/mrm.27420} 
directly 
interpolate missing frequency components in k-space. These methods better 
preserve acquired measurements but often fail to fully exploit spatial 
correlations available in the image domain.

To overcome the limitations of single-domain learning, cross-domain methods 
were introduced to jointly exploit complementary information in both k-space 
and image space. \textbf{AUTOMAP}\cite{Zhu2018} demonstrated early evidence of 
this idea by learning a direct mapping from sensor-domain measurements to image 
space. Later, \textbf{KIKI-Net}\cite{https://doi.org/10.1002/mrm.27201} introduced 
interleaved k-space and image-domain convolutional modules which
integrated frequency and spatial priors through multi-stage supervision.\textbf{DuDoRNet}\cite{9156506} further improved cross-domain reconstruction through image-domain and k-space-domain restoration with interleaved data consistency updates, enabling progressive refinement in both domains simultaneously. These approaches 
showed that alternating between both domains can improve reconstruction quality 
beyond single-domain learning.\\

While purely data-driven models achieve impressive performance, they often lack 
interpretability and may violate physical consistency. To address this, 
model-driven approaches integrate deep learning with classical optimization 
frameworks. Unrolled networks such as \textbf{ADMM-Net}\cite{NIPS2016_1679091c}, 
along with variants based on \textbf{PDHG}\cite{9ea10c47ba7844be8eed9a0956a5946d} 
and \textbf{ISTA}\cite{zhang2018ista}, transform iterative optimization steps into trainable architectures. \textbf{TransMed}\cite{diagnostics11081384} demonstrated the effectiveness of hybrid CNN-transformer architectures for multi-modal medical imaging by combining convolutional feature extraction with transformer-based long-range dependency modeling across modalities. Gradient-based unrolling was later extended by Variational Networks 
(\textbf{VN})\cite{https://doi.org/10.1002/mrm.26977}, which learn regularization 
functions and data consistency parameters directly from data.\\

In parallel, generative models including Generative Adversarial Networks GANs, diffusion models and flow-matching have emerged as promising alternatives for MRI reconstruction. \textbf{DiffuseRecon}\cite{10635308} introducded stochastic diffusion-based sampling guided by observed k-space measurements, enabling reconstruction across varying acceleration factors without retraining. \textbf{HFS-SDE}\cite{10385176} also improved the diffusion-based MRI reconstruction by limiting the stochastic diffusion process to high frequency space, which resulted in better preservation of fine image details while reducing instability and sampling complexity during the reverse diffusion. Recent flow-matching method, \textbf{SMSflow}\cite{WANG2025102593} reduce inference complexity even further by formulating the reconstruction as a continuous ODE-based transport between the undersampled and fully reconstructed images allowing for much faster sampling than conventional diffusion models. Though these methods achieve impressive 
reconstruction quality, they typically require large-scale training data and significant computational resources.\\

Although significant progress has been made in the existing MRI reconstruction methods, there still exist certain difficulties in preserving the fine details, maintaining the global structural consistency, and robustness for high acceleration factors, primarily due to the fact that the conventional convolutional networks usually adopt pooling operations that discard important high-frequency information. To overcome these limitations, we propose a Wavelet-Based Variational Network, which incorporates multi-scale frequency decomposition into an unrolled reconstruction framework, and replaces the standard pooling with Discrete Wavelet Transform (DWT)-based downsampling, which preserves both structural and detail information and physics-guided data consistency and sensitivity estimation further improve reconstruction fidelity and robustness for accelerated MRI.

\section{Proposed Method}

\subsection{Problem Formulation}

In magnetic resonance imaging (MRI), the measured data are acquired in the spatial-frequency domain, commonly referred to as \emph{k-space}. For a multi-coil acquisition with $N_c$ receiver coils, each coil observes the underlying complex-valued image $x \in \mathbb{C}^{H \times W}$ modulated by its spatial sensitivity profile. The measurement from the $i$-th coil can be written as

\begin{align}
k_i = \mathcal{F}(S_i x) + \epsilon_i, \qquad i = 1,2,\dots,N_c
\tag{1}
\end{align}

where $S_i$ denotes the coil sensitivity map of the $i$-th coil, $\mathcal{F}(\cdot)$ is the 2D Fourier transform, and $\epsilon_i$ represents measurement noise. The sensitivity maps are typically normalized such that

\begin{align}
\sum_{i=1}^{N_c} S_i^{*} S_i = I
\tag{2}
\end{align}

where $(\cdot)^*$ denotes complex conjugate transpose. Stacking all coil measurements gives the complete fully-sampled multi-coil k-space data $k$.

Although fully-sampled acquisition enables high-fidelity reconstruction, MRI scan time is directly related to the number of acquired k-space lines. Dense sampling therefore leads to long acquisition time, increased patient discomfort, and susceptibility to motion artifacts. To accelerate the scan, only a subset of k-space locations is sampled using a binary undersampling mask $M$, yielding

\begin{align}
\tilde{k}_i = M \odot k_i, \qquad i = 1,2,\dots,N_c
\tag{3}
\end{align}

where $\odot$ denotes element-wise multiplication and $\tilde{k}_i$ is the acquired undersampled measurement. However, direct inverse Fourier transformation of $\tilde{k}_i$ violates Nyquist sampling conditions and produces severe aliasing artifacts in image space.

Therefore, accelerated MRI reconstruction is formulated as a regularized inverse problem that seeks to recover the latent image $x$ from undersampled measurements:

\begin{align}
\hat{x} = \arg\min_x \frac{1}{2}\left\| \mathcal{T}(x)-\tilde{k} \right\|_2^2 + \lambda \mathcal{R}(x)
\tag{4}
\end{align}

where $\tilde{k}$ denotes the stacked undersampled multi-coil measurements, $\mathcal{T} = M \mathcal{F} S$ is the forward encoding operator consisting of coil sensitivity modulation, Fourier transform, and undersampling, $\mathcal{R}(x)$ is a learned or handcrafted image prior, and $\lambda$ controls the regularization strength. The first term enforces measurement consistency, while the second promotes anatomically plausible reconstructions.

Unrolling an iterative gradient-descent solver for Eq.~(4) gives the image-domain update rule

\begin{align}
x^{t+1} = x^t - \eta^t \mathcal{T}^{*}\!\left(\mathcal{T}(x^t)-\tilde{k}\right)
+ \lambda \mathcal{G}(x^t)
\tag{5}
\end{align}

where $\eta^t$ is the learnable step size at iteration $t$, $\mathcal{T}^{*}$ is the adjoint operator, and $\mathcal{G}(\cdot)$ denotes the gradient of the learned regularizer.

To make the transition to k-space explicit, we expand the forward and adjoint operators. Since $\mathcal{T} = M \mathcal{F} S$, its adjoint is given by

\begin{align}
\mathcal{T}^* = S^* \mathcal{F}^{-1} M
\tag{5a}
\end{align}

Substituting this into Eq.~(5), the update becomes

\begin{align}
x^{t+1} = x^t - \eta^t S^* \mathcal{F}^{-1} \big[ M \odot (\mathcal{F} S x^t - \tilde{k}) \big]
+ \lambda \mathcal{G}(x^t)
\tag{5b}
\end{align}

We now define the k-space representation of the current estimate as

\begin{align}
K^t = \mathcal{F}(S x^t)
\tag{5c}
\end{align}



To obtain an update directly in k-space, we substitute Eq.~(5c) into Eq.~(5b) and apply the operator $\mathcal{F} S$ to both sides of Eq.~(5b), which yields

\begin{align}
K^{t+1} = K^t - \eta^t \mathcal{F} S S^* \mathcal{F}^{-1} \big[ M \odot (K^t - \tilde{k}) \big]
+ \lambda \,\mathcal{F} S \mathcal{G}(x^t)
\tag{5d}
\end{align}

This expression corresponds to the exact k-space update implied by the image-domain optimization. In practice, following common assumptions in parallel MRI, the composite operator $\mathcal{F} S S^* \mathcal{F}^{-1}$ is approximated as an identity mapping, leveraging the normalization condition in Eq.~(2). Furthermore, the learned regularizer is implemented directly in k-space as $\mathcal{G}(K^t)$. Under these approximations, the update simplifies to

\begin{align}
K^{t+1} =
\underbrace{K^t - \eta^t \big[M \odot (K^t-\tilde{k})\big]}_{\text{Data Consistency Block}}
+ \underbrace{\lambda \mathcal{G}(K^t)}_{\text{W-UNet}}
\tag{6}
\end{align}

where $K^t$ is the estimated multi-coil k-space at cascade $t$. The first term enforces data consistency by preserving acquired samples, while the second term corresponds to the proposed W-UNet refinement module. Although the final reconstruction is evaluated in image space, this formulation aligns the optimization more closely with the physics of MRI acquisition, enabling effective recovery of missing frequency components while leveraging learned priors.

\subsection{Proposed Variational Wavelet Network}

\subsubsection*{\textbf{Overall Architecture}}

The proposed framework follows an unrolled variational reconstruction paradigm operating in k-space. 
Given undersampled multi-coil k-space measurements, the model first estimates coil sensitivity maps 
using a dedicated Sensitivity Map Estimation (SME) block. These estimated sensitivity maps are then 
used within an iterative reconstruction process consisting of multiple cascaded refinement steps. 

\begin{figure*}[ht]
    \centering
    \includegraphics[width=\textwidth]{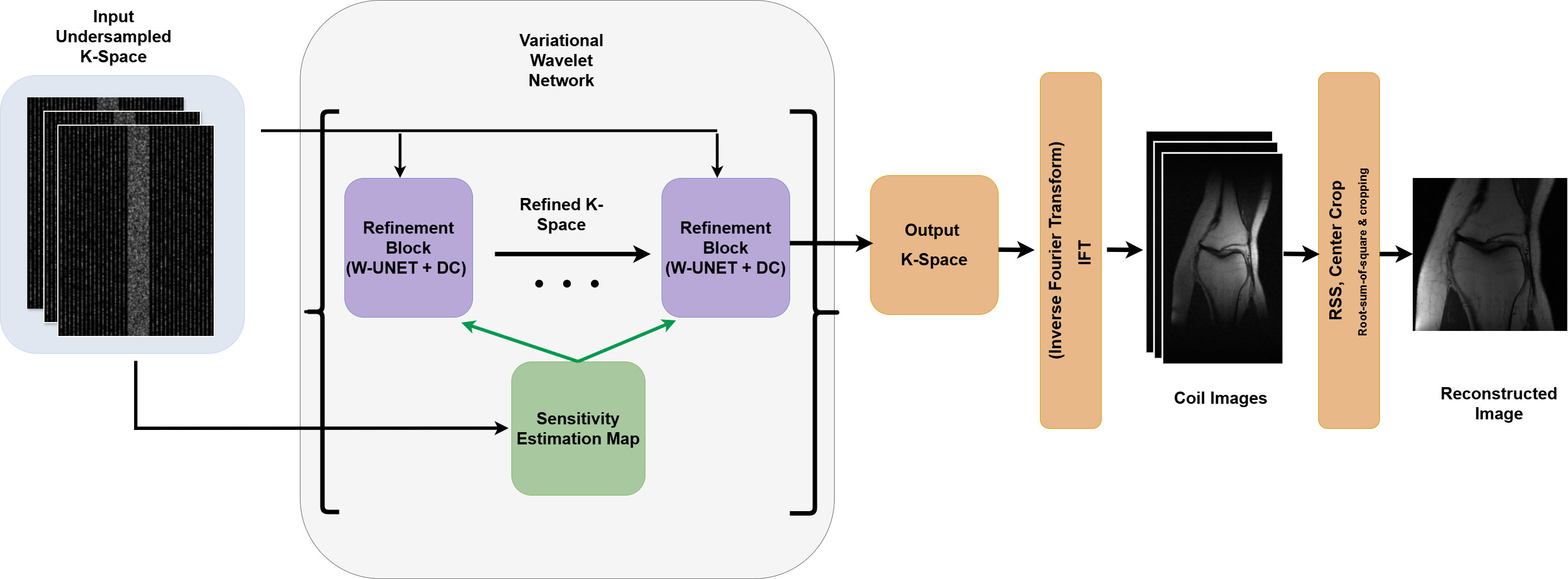}
    \caption{Overall Model Architecture}
    \label{fig:model}
\end{figure*}

Each cascade comprises two key components: a Refinement Block (RB), which performs learned 
regularization in the image domain using a Wavelet-Based U-Net (W-UNet), and a Data Consistency (DC) 
block, which enforces fidelity to the acquired k-space measurements. The cascaded structure enables 
progressive refinement of the reconstruction, effectively reducing artifacts introduced by 
undersampling. The overall pipeline of the proposed model is illustrated in Fig.~\ref{fig:model}.

\subsubsection*{\textbf{Sensitivity Map Estimation Block (SME)}}

Accurate coil sensitivity maps are essential for combining multi-coil MRI measurements into a single coherent image. 
In practice, accelerated MRI acquisition is achieved via \textit{retrospective undersampling}, where a binary sampling 
mask $\mathbf{M}$ is applied to fully-sampled k-space data $\mathbf{K}$:
\begin{align}
    \mathbf{K}_{\text{us}} = \mathbf{M} \odot \mathbf{K},
\end{align}
where $\odot$ denotes element-wise multiplication. The mask $\mathbf{M}$ is designed to retain a fully-sampled 
central region of k-space, known as the \textit{Auto-Calibration Signal (ACS)} region, while sparsely sampling the 
high-frequency components. This sampling strategy mimics practical acquisition constraints and ensures that 
low-frequency structural information is preserved, which is critical for calibration-based reconstruction methods 
such as GRAPPA\cite{https://doi.org/10.1002/mrm.10171}

Let $\mathbf{K}_{\text{ACS}} \subset \mathbf{K}_{\text{us}}$ denote the extracted central k-space region containing 
fully-sampled low-frequency lines. This region is used exclusively for sensitivity estimation due to its high 
signal-to-noise ratio and dense sampling, which enables stable estimation of coil profiles. In contrast, the 
remaining undersampled high-frequency regions are not suitable for direct sensitivity estimation due to missing data 
and aliasing artifacts.

The Sensitivity Map Estimation (SME) block operates by first transforming the ACS k-space data into the image domain:
\begin{align}
    \mathbf{I}_{\text{ACS}} = \mathcal{F}^{-1}(\mathbf{K}_{\text{ACS}}),
\end{align}
where $\mathcal{F}^{-1}$ denotes the inverse Fourier transform applied coil-wise. The resulting coil images encode 
spatial sensitivity variations of each receiver coil.

These initial estimates are further refined using a Wavelet-based U-Net (W-UNet), which captures both low-frequency 
smooth variations and high-frequency boundary transitions in the sensitivity profiles. The final sensitivity maps are 
obtained as:
\begin{align}
    \hat{\mathbf{S}} 
    &= \mathcal{N}\!\left(\mathcal{U}_\mathcal{W}\!\left(\mathcal{F}^{-1}(\mathbf{K}_{\text{ACS}})\right)\right), \tag{7}
\end{align}
where:
\begin{itemize}
    \item $\mathbf{K}_{\text{ACS}} \in \mathbb{C}^{N_c \times N_x \times N_y}$: fully-sampled central k-space region (ACS),
    \item $\mathcal{F}^{-1}$: inverse Fourier transform applied per coil,
    \item $\mathcal{U}_\mathcal{W}(\cdot)$: W-UNet for refining coil-wise spatial features,
    \item $\mathcal{N}(\cdot)$: channel-wise normalization enforcing
    \begin{align}
        \sum_{i=1}^{N_c} |\hat{S}_i(\mathbf{r})|^2 = 1, \quad \forall \mathbf{r},
    \end{align}
    which ensures proper energy normalization across coils.
\end{itemize}


By leveraging the multi-scale decomposition of W-UNet, the SME block effectively models both global sensitivity 
patterns (captured in the low-frequency sub-bands) and localized coil variations (captured in high-frequency sub-bands), 
resulting in more accurate and robust sensitivity maps compared to conventional approaches such as RSS-based or 
polynomial smoothing methods.

\begin{figure*}[ht]
    \centering
    \includegraphics[width=\textwidth]{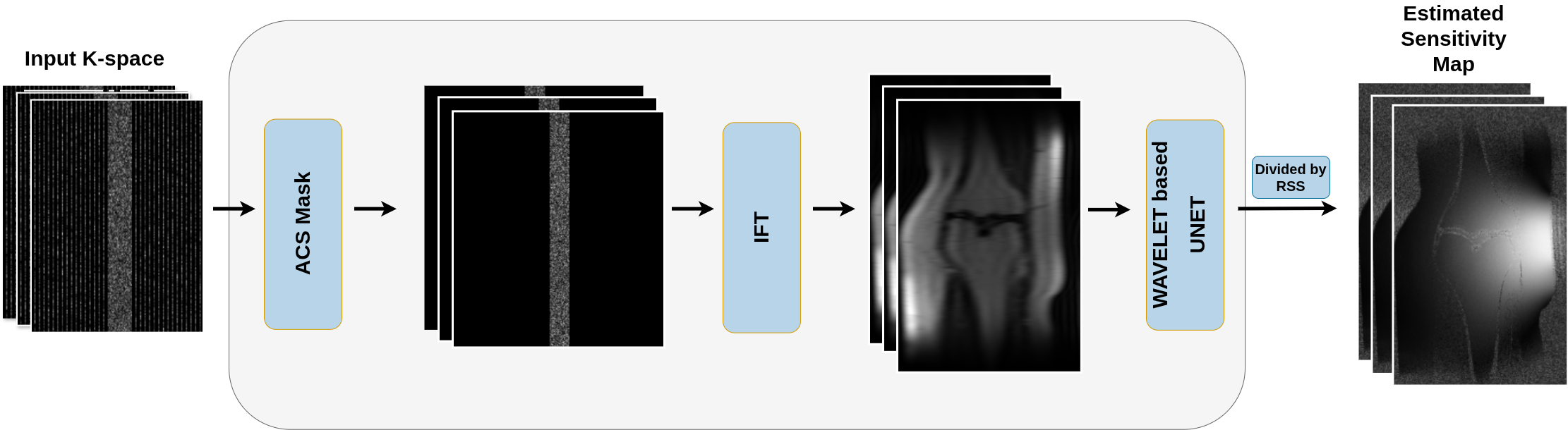}
    \caption{Sensitivity Map Estimation Block}
    \label{fig:sme}
\end{figure*}

\subsubsection*{\textbf{Iterative Refinement via Cascades}}

Following sensitivity estimation, the reconstruction is performed through a sequence of 
iterative refinement steps. Starting from the initial undersampled k-space input 
$\mathbf{K}^0 = \tilde{\mathbf{k}}$, the model applies multiple cascades, each consisting 
of a Refinement Block (RB) followed by a Data Consistency (DC) update. At iteration $t$, 
the k-space is progressively refined to obtain $\mathbf{K}^{t+1}$.

This unrolled formulation enables the network to iteratively improve reconstruction quality 
by alternating between learned regularization (via RB) and physics-based data fidelity 
enforcement (via DC).

\subsubsection*{\textbf{Refinement Block (RB)}}

The Refinement Block (RB) operates in k-space while leveraging image-domain 
processing for learned regularization. Its primary objective is to suppress 
artifacts introduced by undersampling through a sequence of physically grounded 
transformations and learned feature refinement.

At each iteration $t$, the multi-coil k-space input 
$\mathbf{K}^t \in \mathbb{C}^{N_c \times N_x \times N_y}$ is first mapped 
to the image domain using an inverse Fourier transform followed by a 
coil-combination (reduce) operation. The resulting single combined image is 
then processed by the W-UNet. Finally, the refined image is projected back 
to multi-coil k-space via coil expansion and Fourier transform:

\begin{align}
\mathbf{K}^{t}_{\text{ref}} 
&= \mathcal{F} \Big( \mathcal{E} \big( 
\mathcal{U}_{\mathcal{W}} \big( 
\mathcal{R}(\mathcal{F}^{-1}(\mathbf{K}^t), \mathbf{S}) 
\big), \mathbf{S} \big) \Big)
\end{align}

where:
\begin{itemize}
    \item $\mathcal{F}$ and $\mathcal{F}^{-1}$ denote the Fourier and inverse Fourier transforms, respectively,
    \item $\mathcal{R}(\cdot, \mathbf{S})$ is the \textbf{reduce operation}, which combines multi-coil images into a single image using coil sensitivity maps,
    \item $\mathcal{E}(\cdot, \mathbf{S})$ is the \textbf{expand operation}, which projects a single image back to multi-coil space,
    \item $\mathbf{S}$ represents the estimated coil sensitivity maps,
    \item $\mathcal{U}_{\mathcal{W}}$ denotes the W-UNet operating in the image domain.
\end{itemize}

The reduce and expand operations are defined as:

\begin{align}
\mathcal{R}(\mathbf{K}, \mathbf{S}) 
&= \sum_{c=1}^{N_c} \mathcal{F}^{-1}(\mathbf{K}_c) \cdot \mathbf{S}_c^* 
\\
\mathcal{E}(\mathbf{x}, \mathbf{S}) 
&= \mathcal{F} \big( \mathbf{x} \cdot \mathbf{S} \big)
\end{align}

The \textbf{reduce operation} transforms multi-coil k-space into a single 
image by first applying the inverse Fourier transform and then combining 
coil images using the complex conjugate of sensitivity maps. This corresponds 
to a sensitivity-weighted reconstruction (adjoint operation in parallel MRI).

The \textbf{expand operation} performs the inverse process: it multiplies the 
refined image with the coil sensitivity maps to reconstruct per-coil images, 
followed by a Fourier transform to return to k-space.

The image-domain processing via W-UNet enables multi-scale feature extraction 
through wavelet decomposition, allowing the network to effectively capture 
both low-frequency structural information and high-frequency details. This 
leads to improved artifact suppression and better preservation of anatomical 
structures compared to standard CNN-based refinement.

\begin{figure*}[ht]
    \centering
    \includegraphics[width=\textwidth]{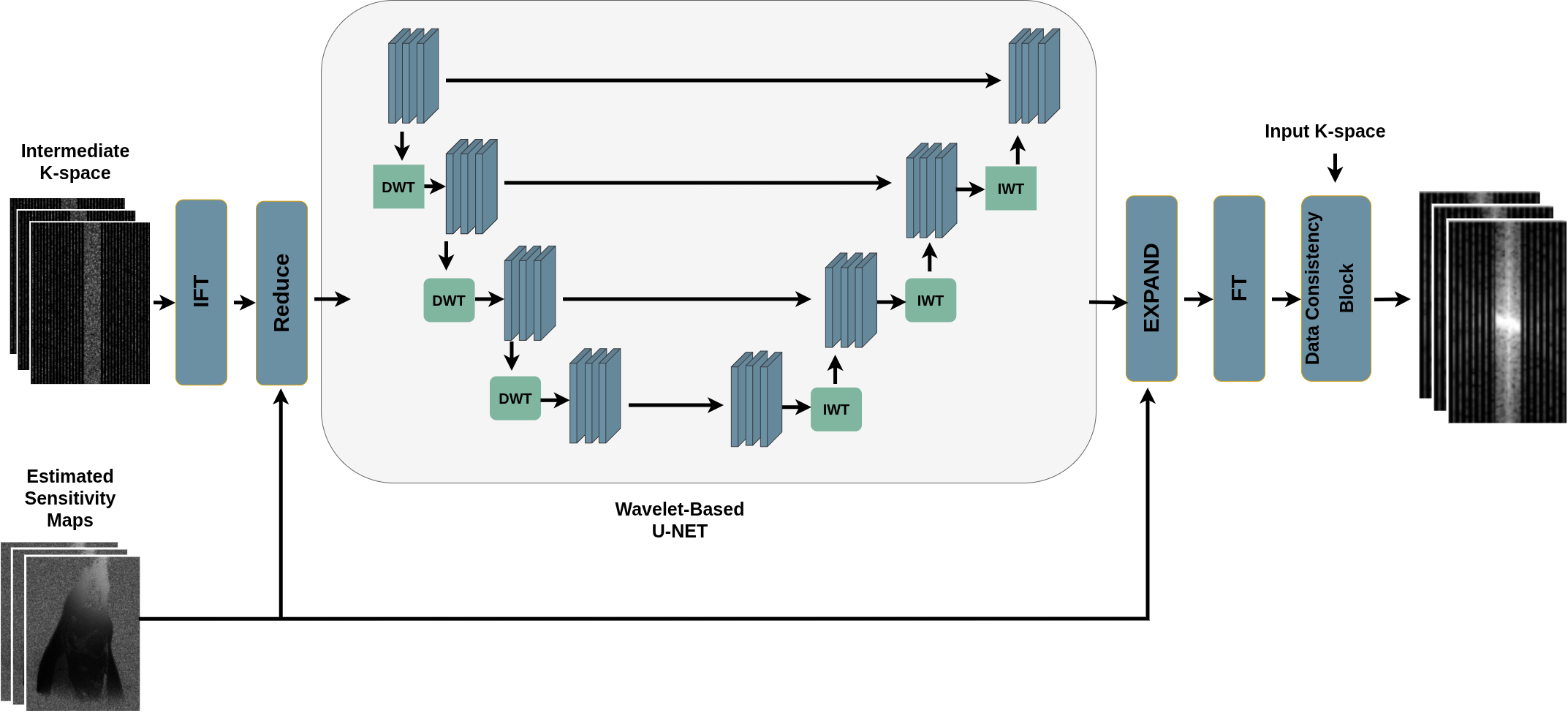}
    \caption{Refinement Block}
    \label{fig:refinement}
\end{figure*}

\subsubsection*{\textbf{Data Consistency (DC) Block}}

The Data Consistency (DC) block enforces agreement between the reconstructed k-space 
and the acquired measurements at each iteration by explicitly correcting the predicted 
values at sampled locations. Let $\mathbf{K}^t$ denote the current k-space estimate 
at iteration $t$, $\tilde{\mathbf{k}}$ the acquired undersampled measurements, and 
$M \in \{0,1\}^{N_x \times N_y}$ the binary undersampling mask, where $M(i,j)=1$ 
indicates sampled locations and $M(i,j)=0$ indicates unsampled locations.

The update is defined as:
\begin{align}
    \mathbf{K}^{t+1} 
    &= \mathbf{K}^{t} - \eta^t\, M\!\left(\mathbf{K}^t - \tilde{\mathbf{k}}\right) \tag{8}
\end{align}
where $\eta^t \in (0,1]$ is a learned step size controlling the strength of the correction.

The update is applied only at sampled locations (i.e., where $M=1$). At these locations, 
the residual error $(\mathbf{K}^t - \tilde{\mathbf{k}})$ is computed and subtracted from 
the current estimate, ensuring that the reconstructed k-space progressively matches the 
true measured data. For unsampled locations ($M=0$), no correction is applied, allowing 
the network to freely predict the missing k-space values. In this way, the model is not 
allowed to alter the reliable measured data, but is instead forced to learn and refine 
only the missing regions of k-space.

When $\eta^t = 1$, the update reduces to hard data consistency:
\[
\mathbf{K}^{t+1} = (1 - M)\odot \mathbf{K}^t + M \odot \tilde{\mathbf{k}},
\]
which directly replaces the sampled k-space values with the acquired measurements.

This iterative correction mechanism improves reconstruction quality by reducing the 
error at sampled locations after each refinement step. As a result, inconsistencies 
introduced by the Refinement Block are progressively eliminated, and the network is 
guided to produce more accurate estimates in the unsampled regions. This leads to 
improved recovery of high-frequency details and suppression of undersampling artifacts.

Although the update is performed in k-space, its effect propagates to the image domain 
through the inverse Fourier transform. Since the measured k-space values are preserved, 
no information is lost at sampled locations. However, strict enforcement (i.e., 
$\eta^t = 1$) may introduce slight inconsistencies in the image domain due to residual 
errors in sensitivity map estimation or model predictions. The learnable step size 
$\eta^t$ alleviates this issue by enabling soft enforcement, balancing measurement 
fidelity with model-based refinement and improving overall reconstruction quality.

\begin{figure}[ht]
    \centering
    \includegraphics[width=\columnwidth]{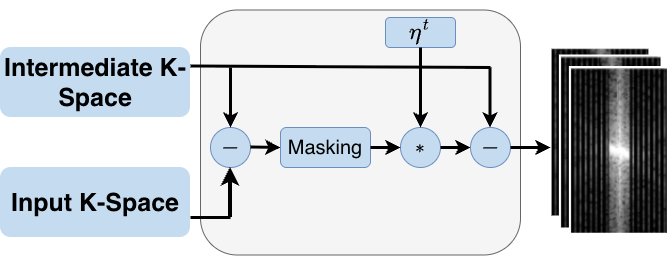}
    \caption{Data Consistency Block}
    \label{fig:dc}
\end{figure}

\subsubsection*{\textbf{Wavelet-Based U-Net (W-UNet)}}
\begin{figure*}[ht]
    \centering
    \includegraphics[width=\textwidth]{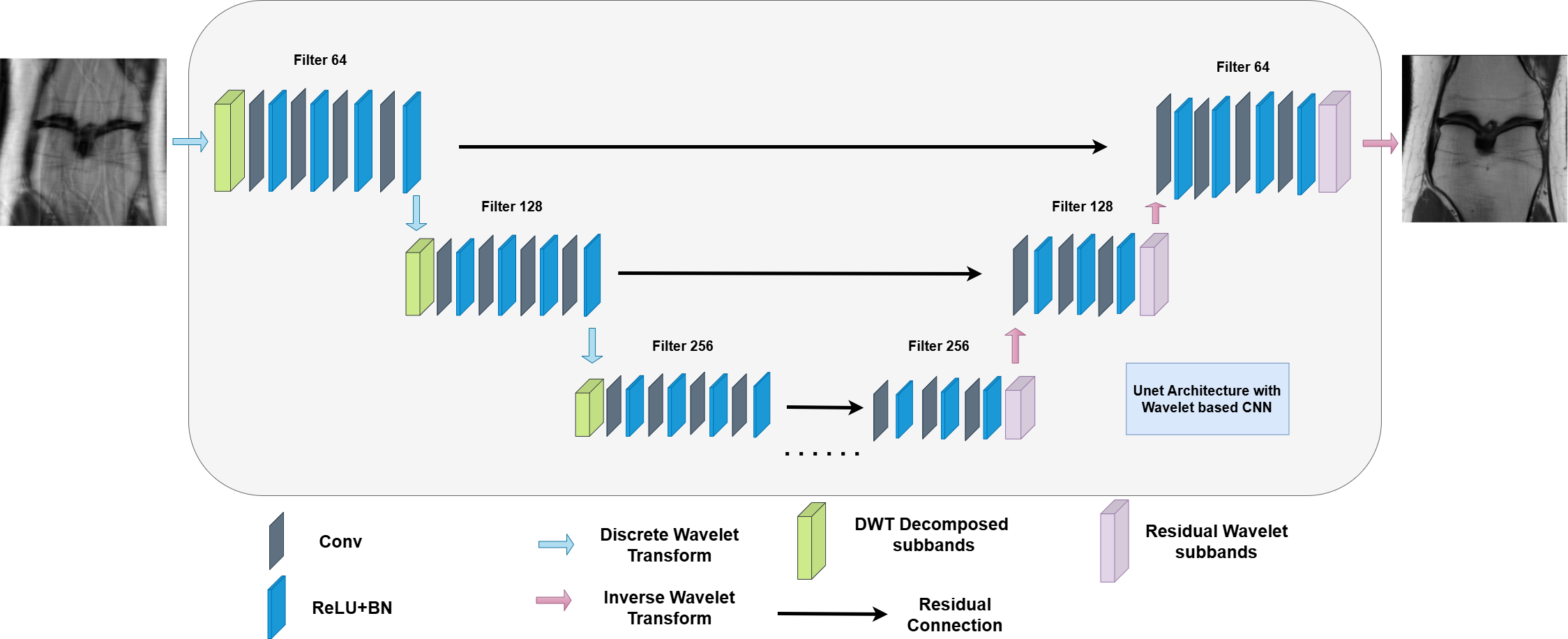}
    \caption{Wavelet-Based U-Net (W-UNet) Architecture}
    \label{fig:wcnn}
\end{figure*}

Standard U-Net architectures employ max-pooling for downsampling, which inherently discards high-frequency components and fine structural details. This limitation is particularly critical in MRI reconstruction, where preserving subtle anatomical features and edge information is essential for diagnostic accuracy. To address this issue, we replace conventional pooling operations with Discrete Wavelet Transform (DWT)-based downsampling, resulting in the proposed Wavelet-Based U-Net (W-UNet), as illustrated in Fig.~\ref{fig:wcnn}. This design enables the network to retain both low and high-frequency information during feature extraction, thereby improving reconstruction fidelity under undersampling conditions.

Given an input feature map $\mathbf{F} \in \mathbb{R}^{C \times H \times W}$, the DWT decomposes it into four sub-bands:
\begin{align}
    \{LL,\; LH,\; HL,\; HH\} &= \mathcal{W}(\mathbf{F}) \tag{4}
\end{align}
where $LL$ represents the low-frequency approximation component, capturing the coarse global structure of the image. The remaining sub-bands encode high-frequency directional details: $LH$ captures horizontal edge information (low-frequency vertically, high-frequency horizontally), $HL$ captures vertical edges (high-frequency vertically, low-frequency horizontally), and $HH$ represents diagonal details and fine textures.

These sub-bands are concatenated channel-wise and passed to subsequent convolutional layers, preserving frequency-domain information while reducing spatial dimensions by a factor of two. This decomposition allows the network to process structural and textural information separately, leading to improved feature representation compared to standard pooling.

In the decoder path, the Inverse Discrete Wavelet Transform (IDWT) replaces transposed convolutions for upsampling:
\begin{align}
    \hat{\mathbf{F}} &= \mathcal{W}^{-1}(LL,\; LH,\; HL,\; HH) \tag{5}
\end{align}
This formulation provides an inherently multi-scale representation by retaining both global and local frequency components across encoder--decoder skip connections. Unlike standard pooling, wavelet-based downsampling is lossless and invertible, ensuring that no information is discarded during the resolution reduction process.

Furthermore, wavelet representations are naturally sparse in the transform domain, which enhances robustness to noise and undersampling artifacts. This property is particularly advantageous in accelerated MRI reconstruction, where missing k-space information must be accurately recovered while preserving fine anatomical details.

\section{Datasets}

Two publicly available benchmark datasets, fastMRI~\cite{Zbontar2018fastMRIAO} and M4Raw~\cite{Lyu2023M4Raw}, are utilized in this study. Both datasets contain raw k-space complex data alongside their corresponding reconstructed images. Experiments are conducted on \textbf{fastMRI Single-Coil Knee}, \textbf{fastMRI Multi-Coil Knee}, and \textbf{M4Raw Multi-Coil Brain} subsets. The volume distribution across training and validation splits is presented in Table~\ref{tab:dataset_info}.

\renewcommand{\arraystretch}{1.5}
\begin{table}[h]
\centering
\resizebox{\linewidth}{!}{
\begin{tabular}{|c|c|c|c|c|}
\hline
\textbf{Dataset} & \textbf{Anatomy} & \textbf{Modality} & \textbf{Train Vol.} & \textbf{Validation Vol.} \\ \hline
\multirow{4}{*}{fastMRI~\cite{Zbontar2018fastMRIAO}} & \multirow{2}{*}{Knee (Single-coil)} & PD & 484 & 100 \\ \cline{3-5} 
 &  & PDFS & 489 & 99 \\ \cline{2-5} 
 & \multirow{2}{*}{Knee (Multi-coil)} & PD & 158 & 29 \\ \cline{3-5} 
 &  & PDFS & 170 & 31 \\ \hline
\multirow{3}{*}{M4Raw~\cite{Lyu2023M4Raw}} & \multirow{3}{*}{Brain (Multi-coil)} & AXT1 & 308 & 76 \\ \cline{3-5} 
 &  & AX FLAIR & 207 & 49 \\ \cline{3-5} 
 &  & AXT2 & 304 & 80 \\ \hline
\end{tabular}
}
\captionsetup{justification=centering, skip=10pt}
\caption{Training \& Validation volumes from fastMRI and M4Raw}
\label{tab:dataset_info}
\end{table}

Each dataset provides three types of data:

\begin{itemize}
    \item \textbf{Raw Multi-Coil K-space Data:} Complex-valued, unprocessed multi-coil k-space acquisitions.
    \item \textbf{Emulated Single-Coil K-space Data:} Multi-coil k-space data converted to single-coil to emulate single-coil acquisition.
    \item \textbf{Ground-Truth Images:} Real-valued spatial images computed from fully sampled k-space data, provided as \textbf{reconstruction\_rss} (multi-coil) and \textbf{reconstruction\_esc} (single-coil).
\end{itemize}

Both datasets adhere to the ISMRMRD vendor-neutral format, stored in \texttt{.h5} container files. Each volume is organized under three keys:

\begin{itemize}
    \item \textbf{ismrmrd\_header:} An XML string containing file metadata.
    \item \textbf{kspace:} Raw k-space data.
    \item \textbf{reconstruction\_rss / reconstruction\_esc:} Ground-truth reconstructed image.
\end{itemize}

The fastMRI knee subset encompasses two acquisition sequences: Proton-Density (\textbf{PD}) and Proton-Density Fat-Suppressed (\textbf{PDFS}), with volumes averaging 36 slices and 15 coils per multi-coil acquisition, for a total of 34,742 slices. The M4Raw brain subset covers three sequences: \textbf{T1}-weighted, \textbf{T2}-weighted, and \textbf{FLAIR}, with each volume comprising 18 slices (4 coils) and a total of 18,432 slices.

\section{Experimental Setup}
\subsection{Data Preprocessing}
The fastMRI and M4Raw datasets provide fully-sampled raw k-space data, which must be retrospectively under-sampled to simulate accelerated MRI acquisition. Under-sampling is applied via a binary mask $M$ as:
\begin{align}
k'_i = M \odot k_i, \qquad i = 1,2,\dots,N_c
\tag{3}
\end{align}
where $K_i$ is the fully-sampled k-space from the $i$-th coil and $K'_i$ is the resulting under-sampled k-space.

The mask is constructed by retaining a central block of ACS (Auto-Calibration Signal) lines and randomly sampling the remaining high-frequency region. For an acceleration factor $\times 4$, ACS lines constitute $8\%$ of total k-space lines; for $\times 2$, this is $11\%$. Experiments are conducted at both acceleration factors.

The fastMRI knee k-space has dimensions $S \times H \times W$ (single-coil) and $S \times C \times H \times W$ (multi-coil), where $S$ = slices, $C$ = 15 coils, $H$ = 640, $W$ = 368; M4Raw brain data has $C$ = 4 coils and $H = W$ = 256. All volumes are centre-cropped to $256 \times 256$ (knee) and $200 \times 200$ (brain) prior to training.

\begin{figure}[h]
\centering
\includegraphics[width=\columnwidth]{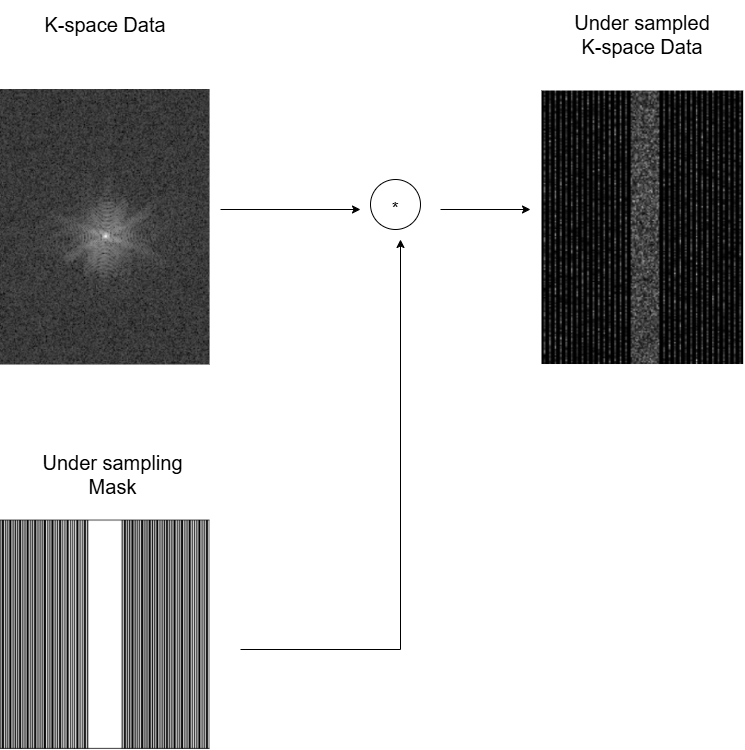}
\caption{Illustration of the k-space undersampling process applied during preprocessing.}
\end{figure}



\subsection{Evaluation Metrics}
Reconstruction quality is assessed using PSNR and SSIM, computed on the validation split. PSNR measures signal fidelity in the logarithmic scale (dB):
\begin{equation*}
    \text{PSNR}(\hat{y}, y) = 10\log_{10}\!\left[\frac{\max(y)^2}{\text{MSE}(\hat{y},y)}\right]
\end{equation*}
SSIM captures perceptual similarity via local statistics:
\begin{equation*}
    \text{SSIM}(x,y) = \frac{(2\mu_x\mu_y+c_1)(2\sigma_{xy}+c_2)}{(\mu_x^2+\mu_y^2+c_1)(\sigma_x^2+\sigma_y^2+c_2)}
\end{equation*}
where $\mu_x, \mu_y$ are local means; $\sigma_x^2, \sigma_y^2$ are variances; $\sigma_{xy}$ is cross-covariance; $c_1, c_2$ are stabilisation constants.

We evaluate our model on the validation split based on two most used metric in reconstruction task - Structural Similarity (SSIM) \& Peak-Signal-to-Noise ratio (PSNR).

\section{Results and Discussion}
The proposed VarNet + W-UNET model consistently outperforms traditional CNN-based methods and state-of-the-art (SOTA) approaches across multiple datasets and acceleration factors. All models were trained with a batch size of one volume per iteration using the Adam optimizer ($\eta = 5 \times 10^{-4}$) and a learning rate scheduler applied every 10 epochs. Due to GPU memory constraints, gradient accumulation was employed every 2 slices for multi-coil data and every 8 slices for single-coil data. Experiments were conducted on an \textbf{NVIDIA Tesla P100} (16\,GB) GPU via the Kaggle platform.



\subsection{Dataset: fastMRI}
\subsubsection{Ours vs. Traditional CNN}
\leavevmode
\begin{table}[H]
    \centering
    \begin{tabular}{|l|c|c|c|}
        \hline
        Model & Modality & \multicolumn{2}{c|}{Metric (*AF = 4)} \\
        \cline{3-4}
         & & SSIM & PSNR(dB) \\
        \hline
        UNET & \multirow{4}{*}{PDFS} & 0.458 & 24.67 \\
        \cline{1-1} \cline{3-4}
        TransUnet & & 0.4883 & 25.11 \\
        \cline{1-1} \cline{3-4}
        W-UNET & & 0.5377 & 28.22 \\
        \cline{1-1} \cline{3-4}
        \textbf{VarNet + W-UNET (ours)} & & \textbf{0.6627} & \textbf{30.96} \\
        \hline
        UNET & \multirow{4}{*} {PD} & 0.673 & 24.924 \\
        \cline{1-1} \cline{3-4}
        TransUnet &  & 0.7043 & 25.37 \\
        \cline{1-1} \cline{3-4}
        W-UNET & & 0.7561 & 28.468 \\
        \cline{1-1} \cline{3-4}
        \textbf{VarNet + W-UNET (ours)} & & \textbf{0.85} & \textbf{31.22} \\
        \hline
        \end{tabular}
    \captionsetup{justification=centering, skip=10pt}
    \caption{Comparison of different methods with SSIM and PSNR metrics at Acceleration Factor(AF) = 4}
    \label{tab:method-comparison}
\end{table} 

\subsubsection{Ours vs. SOTA}
\leavevmode
\begin{table}[H]
    \centering
    \begin{tabular}{|l|c|c|c|}
        \hline
        \multirow{2}{*}{Method} & \multirow{2}{*}{Modality} & \multicolumn{2}{c|}{Metric (AF = 4)} \\
        \cline{3-4}
        & & SSIM & PSNR(dB) \\
        \hline
        CS~\cite{article} & \multirow{13}{*} {PDFS} & 0.5736 & 29.54 \\
        \cline{1-1} \cline{3-4}
        LORAKS~\cite{6678771} & & 0.543 & 25.8 \\
        \cline{1-1} \cline{3-4}
        UNET~\cite{Ronneberger2015UNetCN} & & 0.578 & 27.8 \\
        \cline{1-1} \cline{3-4}
        OUCR~\cite{10.1007/978-3-030-87231-1_2} & & 0.602 & 28.5 \\
        \cline{1-1} \cline{3-4}
        Diffrecon~\cite{10.1007/978-3-031-16446-0_59} & & 0.608 & 28.6 \\
        \cline{1-1} \cline{3-4}
        TC-DiffRecon~\cite{10635308} & & 0.503 & 26.5 \\
        \cline{1-1} \cline{3-4}
        HFS-SDE~\cite{10385176} & & 0.503 & 26.5 \\
        \cline{1-1} \cline{3-4}
        SMSflow~\cite{WANG2025102593} & & 0.605 & 28.8 \\
        \cline{1-1} \cline{3-4}
        Transmed~\cite{diagnostics11081384} & & 0.62 & 28.8 \\
        \cline{1-1} \cline{3-4}
        rsGAN~\cite{Dar2020} & & 0.628 & 30.1 \\
        \cline{1-1} \cline{3-4}
        MTrans~\cite{9796552} & & 0.632 & 30.5 \\
        \cline{1-1} \cline{3-4}
        \textbf{VarNet + W-UNET (ours)} &  & \textbf{0.663} & \textbf{30.96} \\
        
        \hline
        TV~\cite{rudin1992nonlinear} & \multirow{5}{*} {PD} & 0.636 & 27.13 \\
        \cline{1-1} \cline{3-4}
        DuDoRNet~\cite{9156506} & & 0.793 & 30.42 \\
        \cline{1-1} \cline{3-4}
        Score-based Diff. Model~\cite{CHUNG2022102479} & & 0.812 & 31.95 \\
        \cline{1-1} \cline{3-4}
         DiffINR~\cite{CHU2025103398} & & 0.7475 & 30.944 \\
        \cline{1-1} \cline{3-4}
        Kronecker U-Net~\cite{10.1007/978-3-032-09513-8_10} & & 0.762 & 29.78 \\
        \cline{1-1} \cline{3-4}
        \textbf{VarNet + W-UNET (ours)} &  & \textbf{0.85} & \textbf{31.22} \\
        \hline
    \end{tabular}
    \caption{fastMRI: Knee Single-Coil}
    \label{tab:fastmri-knee-single}
\end{table}

\begin{table}[H]
    \centering
    \begin{tabular}{|l|c|c|}
        \hline
        \multirow{2}{*}{Method} & \multicolumn{2}{c|}{Metric (AF = 4)} \\
        \cline{2-3}
        & SSIM & PSNR(dB) \\
        \hline
        TV~\cite{rudin1992nonlinear} & 0.693 & 32.10 \\
        \hline
        VarNet~\cite{https://doi.org/10.1002/mrm.26977} & 0.818 & 32.97 \\
        \hline
        Score-based Diff. Model~\cite{CHUNG2022102479} & 0.857 & 33.25 \\
        \hline
        LMO~\cite{11092909} & 0.832 & 28.65 \\
        \hline
        \textbf{VarNet + W-UNET (ours)} & \textbf{0.903} & \textbf{34.72} \\
        \hline      
    \end{tabular}
    \caption{fastMRI: Knee Multi-Coil}
    \label{tab:fastmri-knee-multi}
\end{table}

\subsection{Dataset: M4Raw}
\begin{table}[H]
    \centering
    \begin{tabular}{|l|c|c|c|}
        \hline
        \multirow{2}{*}{Method} & \multirow{2}{*}{Modality} & \multicolumn{2}{c|}{Metric (AF = 2)} \\
        \cline{3-4}
        & & SSIM & PSNR(dB) \\
        \hline
        GRAPPA~\cite{https://doi.org/10.1002/mrm.10171} &  \multirow{3}{*} {AX FLAIR} & 0.708 & 28.52 \\
        \cline{1-1} \cline{3-4}
        VarNet~\cite{https://doi.org/10.1002/mrm.26977} & & 0.851 & 34.21 \\
        \cline{1-1} \cline{3-4}
        \textbf{VarNet + W-UNET (ours)} &  & \textbf{0.857} & \textbf{32.47} \\
        \hline
        GRAPPA~\cite{https://doi.org/10.1002/mrm.10171} &  \multirow{3}{*} {AX T2} & 0.72 & 28.70 \\
        \cline{1-1} \cline{3-4}
        VarNet~\cite{https://doi.org/10.1002/mrm.26977} & & 0.843 & 33.87 \\
        \cline{1-1} \cline{3-4}
        \textbf{VarNet + W-UNET (ours)} &  & \textbf{0.862} & \textbf{32.60} \\
        \hline
        GRAPPA~\cite{https://doi.org/10.1002/mrm.10171} &  \multirow{3}{*} {AX T1} & 0.743 & 30.71 \\
        \cline{1-1} \cline{3-4}
        VarNet~\cite{https://doi.org/10.1002/mrm.26977} & & 0.888 & 35.71 \\
        \cline{1-1} \cline{3-4}
        \textbf{VarNet + W-UNET (ours)} &  & \textbf{0.894} & \textbf{34.88} \\
        \hline        
    \end{tabular}
    \caption{M4Raw: Brain Multi-Coil}
    \label{tab:M4Raw-comparison}
\end{table}

The experiment results indicate the superiority of \textbf{VarNet + W-UNET} for rapid MRI reconstruction on different datasets. Our method consistently performs better than baseline CNN-based methods and state-of-the-art methods with higher SSIM and PSNR scores. Structural detail preservation and noise removal are enhanced using wavelet-based CNNs, leading to higher quality images. These findings indicate the efficacy of our method in accelerating rapid and precise MRI reconstruction, paving the way for future advances in medical imaging.

\section{Ablation Study}
In this section, we present a series of ablation studies to understand the individual contributions of different blocks and methods to the overall model performance. These experiments help us identify the critical components that enhance the reconstruction quality of our proposed Variational Network with Wavelet-based UNET (W-UNET).

\subsection{Effect of the Data Consistency Block}

The Data Consistency (DC) block plays a crucial role in ensuring that the reconstructed images remain faithful to the acquired k-space data. To evaluate its importance, we conducted experiments with and without the DC block. Table \ref{tab:dc_block} summarizes the results for both knee and brain MRI datasets. 

\begin{table}[h]
    \centering
    \setlength{\tabcolsep}{4pt}
    \begin{tabular}{|c|c|c|c|c|c|}
         \hline
         \multirow{2}{*}{Anatomy} & \multirow{2}{*}{Modality} & \multicolumn{2}{c|}{without DC} & \multicolumn{2}{c|}{with DC} \\
         \cline{3-6}
         & & SSIM & PSNR(dB) & SSIM & PSNR(dB) \\ 
         \hline
         \multirow{2}{*}{Knee single coil} & PD & 0.668 & 26.13 & 0.85 & 31.22 \\
         \cline{2-6}
         & PDFS & 0.575 & 24.19 & 0.663 & 30.96\\
         \hline
         \multirow{2}{*}{knee multi coil} & PD & 0.78 & 27.47 & 0.903 & 34.72 \\ 
         \cline{2-6}
         & PDFS & 0.682 & 28.33 & 0.761 & 31.23 \\
         \hline
         \multirow{3}{*}{Brain multi coil} & AX FLAIR & 0.747 & 28.05 & 0.857 & 32.47 \\
         \cline{2-6}
         & AX T1 & 0.824 & 31.89 & 0.894 & 34.88 \\
         \cline{2-6}
         & AX T2 & 0.83 & 31.18 & 0.862 & 32.60\\
         \hline
    \end{tabular}
    \caption{Performance comparison with and without the Data Consistency (DC) block.}
    \label{tab:dc_block}
\end{table}

As shown in Table \ref{tab:dc_block}, removing the DC block led to a significant drop in performance. For example, in the knee multi-coil dataset, the PSNR decreased from 34.72 to 27.47, and the SSIM dropped from 0.903 to 0.78. Similar trends were observed for the brain multi-coil dataset, confirming the essential role of the DC block in preserving the fidelity of the reconstructed images.

\subsection{Effect of Wavelet-based UNET in the Refinement Block}

The Refinement Block is responsible for improving the quality of the reconstructed images by refining the intermediate outputs. To evaluate the impact of using a Wavelet-based UNET (W-UNET) instead of a traditional CNN in this block, we compared the performance of the two architectures. Table \ref{tab:wcnn_vs_unet} presents the results for both single-coil and multi-coil datasets.

\begin{table}[h]
    \centering
    \begin{tabular}{|c|c|c|c|c|c|}
         \hline
         \multirow{2}{*}{Dataset} & \multirow{2}{*}{Modality} & \multicolumn{2}{c|}{VarNet + UNET} & \multicolumn{2}{c|}{VarNet + W-UNET} \\
         \cline{3-6}
         & & SSIM & PSNR(dB) & SSIM & PSNR(dB) \\
         \hline
         \multirow{2}{*}{knee single coil} & PD & 0.802 & 30.03 & 0.85 & 31.22 \\
         \cline{2-6}
         & PDFS & 0.654 & 28.17 & 0.6627 & 30.96 \\
         \hline
         \multirow{2}{*}{knee multi coil} & PD & 0.887 & 33.22 & 0.903 & 34.72 \\
         \cline{2-6}
         & PDFS & 0.809 & 32.79 & 0.823 & 33.97  \\
         \hline
         \multirow{3}{*}{Brain multi coil} & AX FLAIR & 0.822 & 31.75 & 0.857 & 32.47 \\
         \cline{2-6}
         & AX T1 & 0.859 & 33.64 & 0.894 & 34.88 \\
         \cline{2-6}
         & AX T2 & 0.827 & 31.14 & 0.862 & 32.6\\
         \hline
    \end{tabular}
    \caption{Performance comparison between UNET and Wavelet-based UNET (W-UNET) in the Refinement Block.}
    \label{tab:wcnn_vs_unet}
\end{table}

From Table \ref{tab:wcnn_vs_unet}, the W-UNET consistently outperformed the traditional UNet across all datasets. For example, in the knee multi-coil dataset, W-UNET achieved an SSIM of \textbf{0.903} and PSNR of \textbf{34.72}, compared to \textbf{0.887} and \textbf{33.22} for UNet. Similarly, in the brain multi-coil AX T1 modality, W-UNET improved SSIM from \textbf{0.859} to \textbf{0.894} and PSNR from \textbf{33.64} to \textbf{34.88}, demonstrating its effectiveness in capturing high-frequency details and enhancing reconstruction quality.

\subsection{Summary of Findings}

The ablation studies demonstrate the critical contributions of the Data Consistency block and the Wavelet-based UNET in the Refinement Block to the overall performance of the proposed model. The DC block ensures fidelity to the acquired k-space data, while the W-UNET enhances the reconstruction quality by effectively capturing high-frequency details.

\begin{figure*}[ht]
	\centering
	\includegraphics[width=\textwidth]{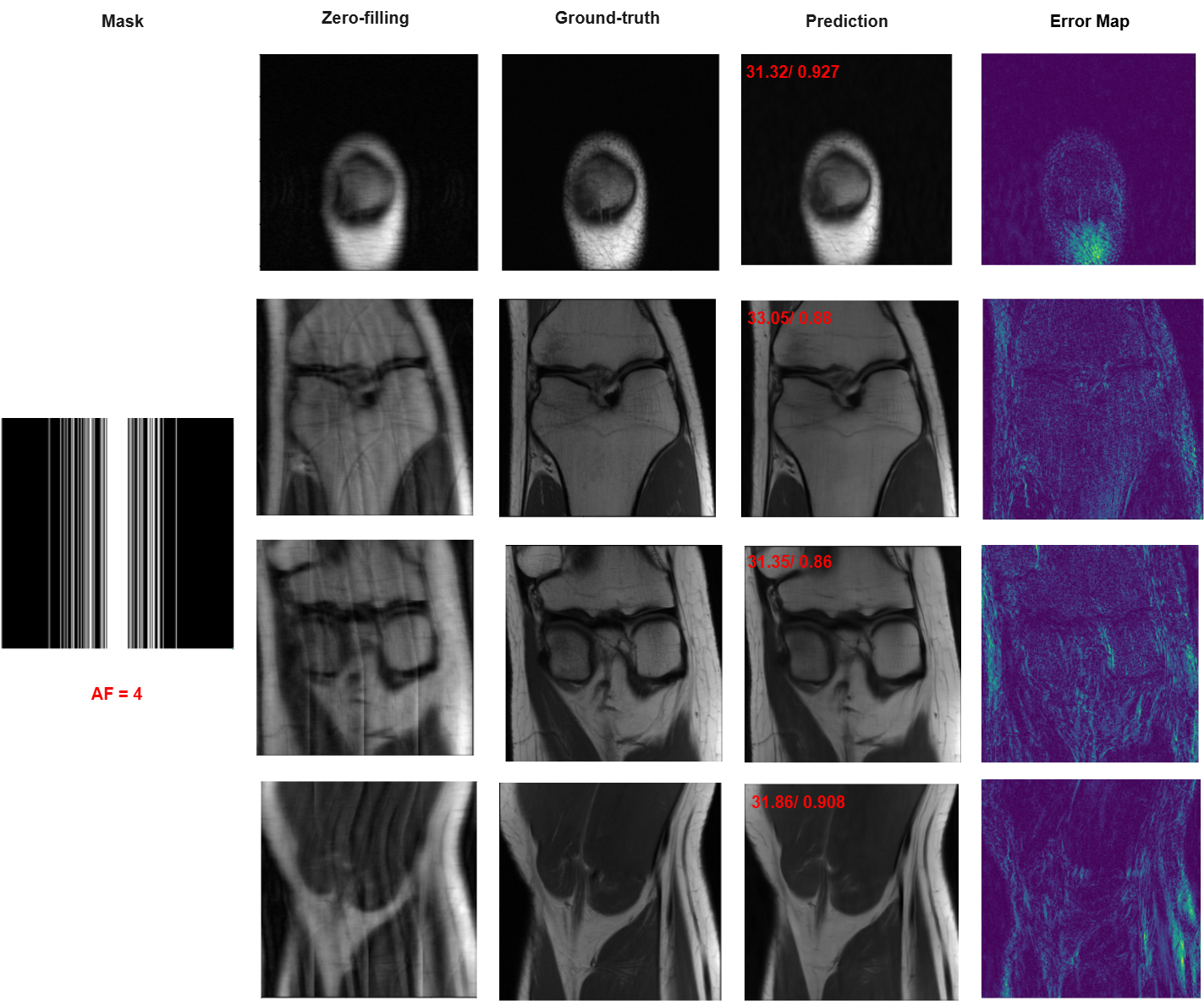}
	\caption{Illustrative Comparison of fastMRI knee Single-Coil MRI Reconstruction (x4 Acceleration Factor)}
	\label{•}
\end{figure*}

\begin{figure*}[ht]
	\centering
	\includegraphics[width=\textwidth]{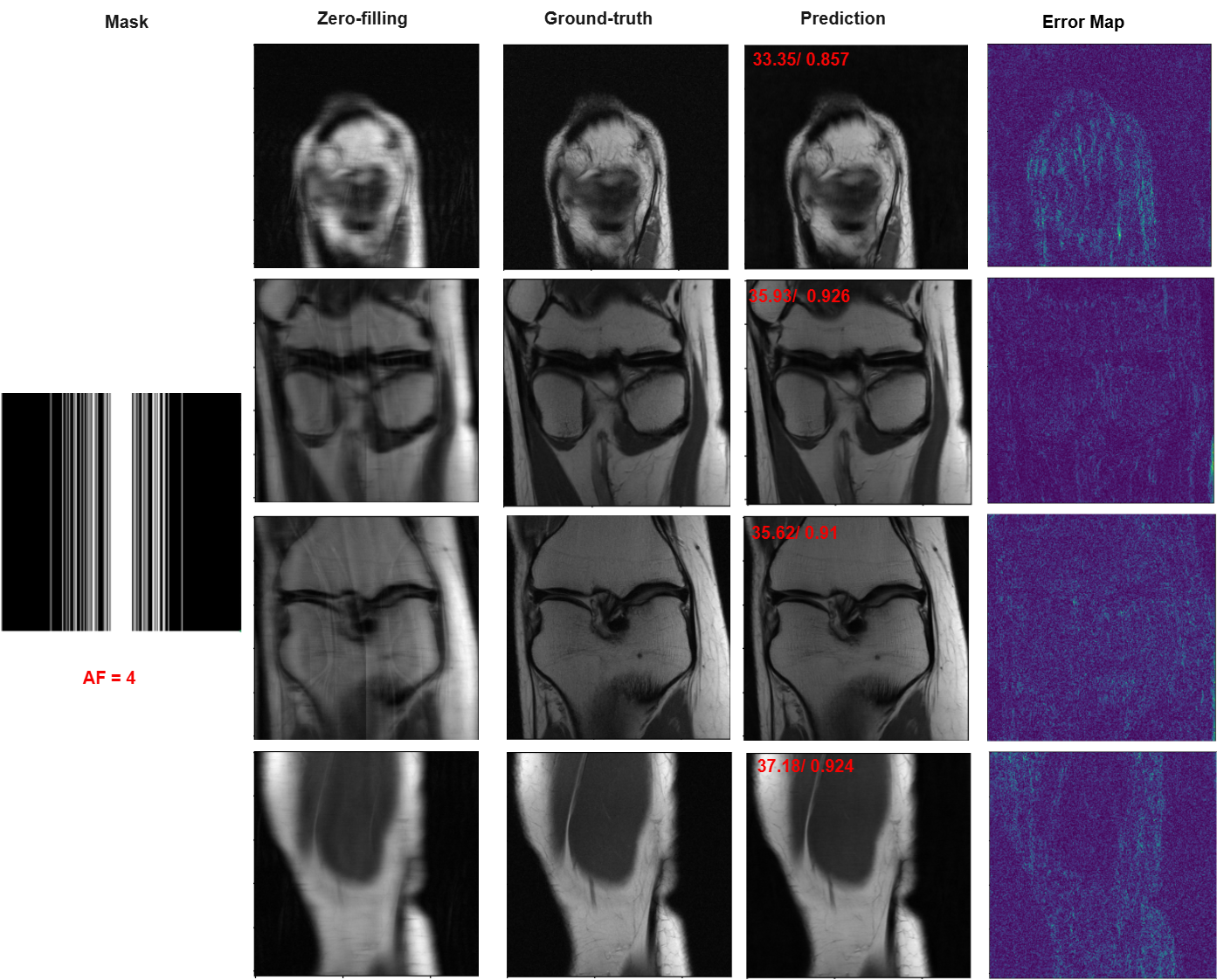}
	\caption{Illustrative Comparison of fastMRI knee Multi-Coil MRI Reconstruction (x4 Acceleration Factor)}
	\label{•}
\end{figure*}

\begin{figure*}[ht]
	\centering
	\includegraphics[width=\textwidth]{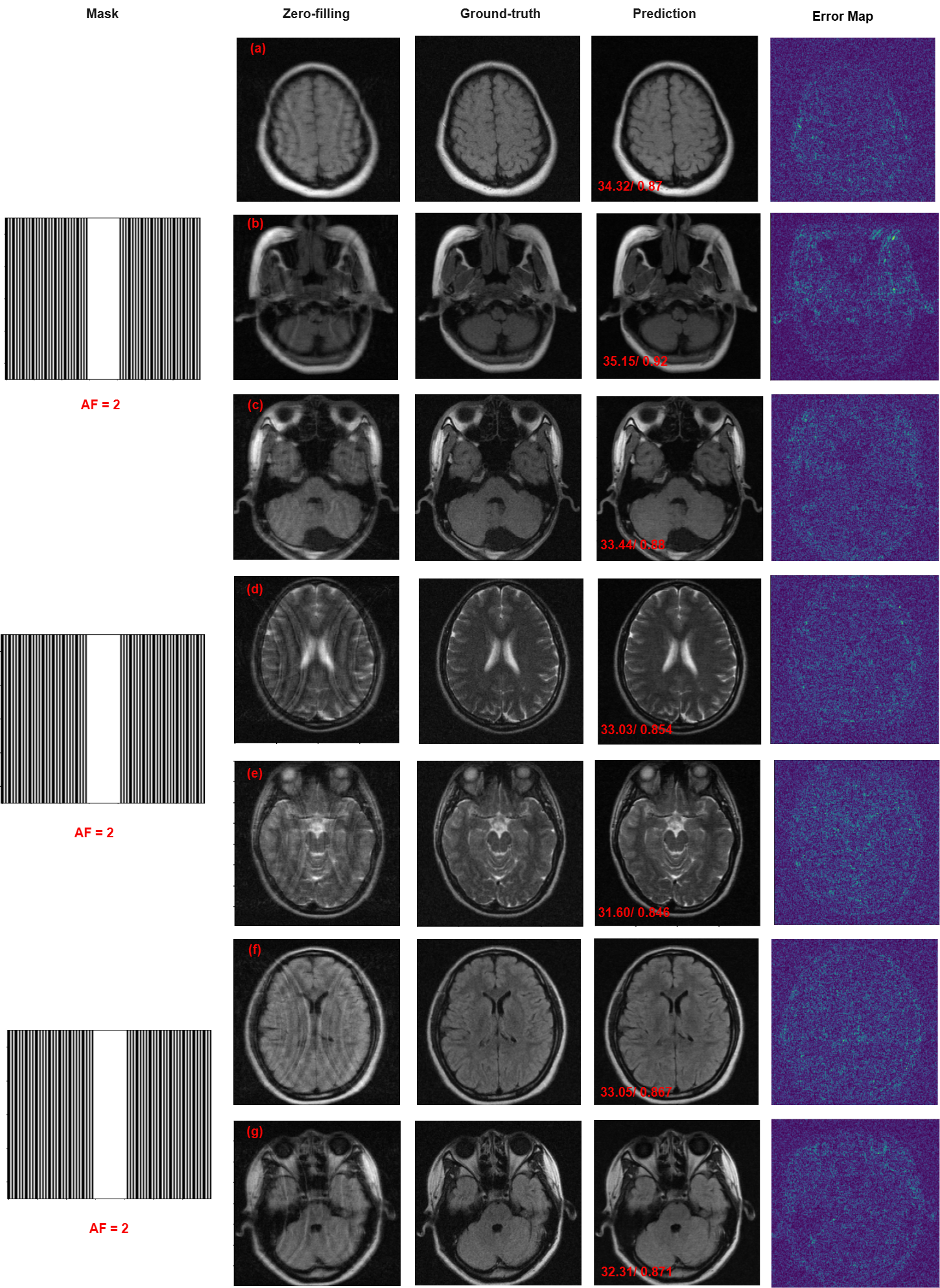}
	\caption{Illustrative Comparison of M4Raw Brain Multi-Coil MRI Reconstruction (x2 Acceleration Factor)}
	\label{•}
\end{figure*}

\clearpage
\balance

\bibliographystyle{ieeetr}
\bibliography{references.bib}


\section{Biography Section}
If you have an EPS/PDF photo (graphicx package needed), extra braces are
 needed around the contents of the optional argument to biography to prevent
 the LaTeX parser from getting confused when it sees the complicated
 $\backslash${\tt{includegraphics}} command within an optional argument. (You can create
 your own custom macro containing the $\backslash${\tt{includegraphics}} command to make things
 simpler here.)
 

\bf{If you include a photo:}\vspace{-33pt}
\begin{IEEEbiography}[{\includegraphics[width=1in,height=1.25in,clip,keepaspectratio]{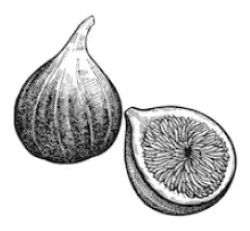}}]{Michael Shell}
Use $\backslash${\tt{begin\{IEEEbiography\}}} and then for the 1st argument use $\backslash${\tt{includegraphics}} to declare and link the author photo.
Use the author name as the 3rd argument followed by the biography text.
\end{IEEEbiography}

\vspace{11pt}

\bf{If you will not include a photo:}\vspace{-33pt}
\begin{IEEEbiographynophoto}{John Doe}
Use $\backslash${\tt{begin\{IEEEbiographynophoto\}}} and the author name as the argument followed by the biography text.
\end{IEEEbiographynophoto}

\vfill

\end{document}